\documentclass[preprint,onefignum,onetabnum]{siamart190516}

\usepackage{lipsum}
\usepackage{amsfonts}
\usepackage{graphicx}
\usepackage{epstopdf}
\usepackage{algorithmic}
\ifpdf
  \DeclareGraphicsExtensions{.eps,.pdf,.png,.jpg}
\else
  \DeclareGraphicsExtensions{.eps}
\fi

\newsiamremark{remark}{Remark}
\newsiamremark{hypothesis}{Hypothesis}
\crefname{hypothesis}{Hypothesis}{Hypotheses}
\newsiamthm{claim}{Claim}

\headers{Refining the Structure of Neural Networks}{R. Yousefzadeh and D. P. O'Leary}

\title{Refining the Structure of Neural Networks Using Matrix Conditioning
}

\author{Roozbeh Yousefzadeh\thanks{Department of Computer Science, University of Maryland, College Park, MD 
  (\email{roozbeh@cs.umd.edu}, \url{http://www.cs.umd.edu/\string~roozbeh/}).}
\and Dianne P. O'Leary\thanks{Department of of Computer Science and Institute for Advanced Computer Studies, University of Maryland, College Park, MD
  (\email{oleary@cs.umd.edu}, \url{http://www.cs.umd.edu/\string~oleary/}).}
}

\usepackage{amsopn}

\ifpdf
\hypersetup{
  pdftitle={Refining the Structure of Neural Networks Using Matrix Conditioning},
  pdfauthor={R. Yousefzadeh and D. P. O'Leary}
}
\fi

\usepackage{xcolor,bm,amsmath,graphicx,comment,bigints}
\usepackage{algorithm,algorithmic}
\graphicspath{{images/}}

\def\bs#1{\boldsymbol{#1}}

\newcommand{\dd}[1]{#1}

\newcommand{\dr}[1]{#1}
\newcommand{\ds}[1]{{#1}}

\newcommand{\olive}[1]{#1}
\newcommand{\rblu}[1]{{#1}}

\usepackage[normalem]{ulem}

\usepackage{booktabs,multirow}

\begin{document}

\maketitle

\begin{abstract}
Deep learning models have proven to be exceptionally useful in performing many machine learning tasks. However, for each new dataset, choosing an effective size and structure of the model can be a time-consuming process of trial and error. While a small network with few neurons might not be able to capture the intricacies of a given task, having too many neurons can lead to overfitting and poor generalization. Here, we propose a practical method that employs matrix conditioning to automatically design the structure of layers of a feed-forward network, by first adjusting the proportion of neurons among the layers of a network and then scaling the size of network up or down. Results on sample image and non-image datasets demonstrate that our method results in small networks with high accuracies. Finally, guided by matrix conditioning, we provide a method to effectively squeeze models that are already trained. Our techniques reduce the human cost of designing deep learning models and can also reduce training time and the expense of using neural networks for applications. 
\end{abstract}




\section{Introduction}


Designing the structure of deep learning models is a delicate and usually time-consuming prerequisite to \dd{using} them for real world applications. The model design approaches in the literature rely on training many models and therefore require \dd{an} extensive amount of computation. Here, we provide a complete set of low-cost computational tools to design \dd{the layers of} a feed-forward neural network from scratch for any dataset, \dd{guided by} matrix conditioning \ds{and partial training}. In this section, we review the literature from different perspectives and relate our approach to previous methods. In \cref{sec_framework}, we define the framework of our methods using a neural network prototype. In \cref{sec_design}, we propose our algorithms to design a network from scratch, and in \cref{sec_squeeze} we describe our method for squeezing networks that are already trained. \Cref{sec_results} contains our numerical results, and finally, conclusions follow in \cref{sec_conclusion}.

\subsection{Model design and its difficulties}

Among the most important decisions to be made in model design is determining an appropriate size for the network. 

The trade-off between the size and accuracy of networks has been studied extensively for benchmark datasets in machine learning, \dd{e.g., }\cite{nowlan1992simplifying,srivastava2014dropout,zhang2016understanding,neyshabur2018the}. Through trial and error, standard models have been developed that can achieve the best accuracies on some of those datasets.
These achievements are impressive, but they do not give us much guidance about how to approach an unfamiliar dataset.

Furthermore, the standard models are often massive and require specialized hardware, which makes them unapproachable for modest real-world tasks. A few studies focus on developing compact models that can achieve acceptable accuracies on standard datasets, e.g., \cite{iandola2016squeezenet,zhou2016less}. Still, there is a great need for systematic and affordable procedures to decide an appropriate number of neurons on each layer of a network for an unfamiliar dataset.

Obtaining a compact model might sometimes come at the cost of losing some accuracy. Nevertheless, that compromise might be justifiable or even necessary in certain applications. The huge computational cost or power consumption for some of the best models is prohibitive for certain computers and applications \cite{canziani2016analysis}, and hence there has been a focus on developing more economical models that maintain acceptable accuracies \cite{denton2014exploiting,han2015learning,howard2017mobilenets}. With that in mind, our focus is not to improve the benchmark accuracies, rather to achieve a modest accuracy with a compact model.

One of the reported advantages of deep learning models is  \dd{the} automatic detection of important features from the raw data, saving the time required for preprocessing and feature selection. That view is not completely correct as we showed in previous work \cite{yousefzadehdebugging}. However, even if  \dd{analysts avoid the cost of} data preprocessing, the structural design of deep models can be very time-consuming. This can become an obstacle in deploying neural networks in mainstream applications, for example problems related to education \cite{jiang2018expert}.

\dd{Alvarez et al.}~\cite{alvarez2016learning} have given a review of earlier approaches to adjusting the size of a neural network. Their method of reducing the size of a neural network is based on adding a penalty term to the loss function in order to detect and remove redundant neurons, while ours  \dd{expands or contracts a network based on} partial matrix decompositions layer by layer. Like their method, we do not need to fully train a network before adjusting its size.

Starting with a large network and adding a regularization term to the loss function of the neural network during training is another common approach to reducing its size. For example, \cite{zhou2016less} imposed sparsity constraints on the dense layers of the standard CNNs and demonstrated that most of the neurons in those models can be eliminated without any degradation of the ``top-1" classification accuracy. Regularization has also been used in other studies, e.g., \cite{murray2015auto} for language models.

Although adding regularization terms in the training process is effective in reducing \dd{over-fitting} for over-sized networks and in identifying redundancies in the standard models, this cannot be considered a direct method to design a neural network from scratch for an unknown dataset. Unlike our \dd{algorithms}, these methods require an over-sized network with high-accuracy to begin with, and their performance depends on specific optimization methods for the training and careful tuning of additional hyperparameters for each dataset. \dd{Our Algorithm \ref{alg_3}} for pruning \dd{trained} over-sized networks does not need to retrain a network from scratch; rather it relies on straightforward and relatively inexpensive row and column elimination from the weight matrices and applying the original training method to \dd{complete the training of} the squeezed network.

\subsection{Model architecture search methods}

\dd{Some resource-intensive methods} consider a pool of candidate models and try to choose the best model, \dd{or  define} networks with a set of parameters \dd{and} then search the space \dd{of}  parameters to find their optimal configuration. 
\dd{Some} earlier \dd{proposals} use statistical methods such as hypothesis testing to find the best models \cite{anders1999model} or genetic algorithms to search the parameter space \cite{stanley2002evolving}. More recently, \cite{zoph2016neural} and \cite{baker2016designing} used reinforcement learning to search the design space, \cite{liu2018progressive} developed a sequential model-based optimization (SMBO) strategy and a surrogate model to guide the search through structure space, \cite{zoph2018learning} used a combination of transfer learning and reinforcement learning, \cite{pham2018efficient} used a method that allowed parameter sharing between the candidate models in order to make the search more efficient, \cite{bender2018understanding} analyzed a class of efficient architecture search methods based on weight sharing, and \cite{hu2019efficient} used a linear regression feature selection algorithm and was successful in finding competitive models using a few GPU days. 

The methods that try to be more efficient risk the possibility of prematurely discarding good candidates that might not appear good in the first stages of training. \cite{cashmanmast} advocates for recycling the training information for the models that are discarded at the initial steps of model search and provides a visual tool to verify assumptions used in the search in order to make the process interactive.

These approaches can be highly effective in finding a good structure for a neural network. However, they can be generally viewed as an automated version of training many networks and finding the best one. Therefore, they are highly resource expensive, \ds{some} \dd{taking even} GPU months or years to find the best neural network architecture for a given task \cite{wistuba2017finding}. This prohibits their use for modest applications with limited computational resources. 

Our \dd{three algoritms have a narrow search strategy and are less costly}. For example, the entire time it takes to train our network for the MNIST dataset on a 2017 Macbook is about two hours. Nevertheless, our goals are similar in the sense that we aim to find the best architecture for a feed-forward neural network. Hence, our methods can be viewed as a low-cost but efficient way to design the structure of networks for mainstream applications in the real-world.

\subsection{\dd{Approaches} based on decomposition of weight matrices}
Here, we consider feed-forward neural networks as a general-purpose machine learning model and develop a training method that can achieve high accuracy by optimizing the number of neurons on each layer of the network, systematically and efficiently.

To achieve our goal, we use the \dd{singular value decomposition (SVD), rank-revealing QR decomposition (RR-QR) \cite{chan1987rank}, or pivoted QR decomposition \cite{golub2012matrix}} of the stacked weight/bias matrices to determine the redundancies in the network and to identify layers that have an excessive number of neurons. 

One of the early \dd{uses} of SVD to \dd{prune}  feed-forward neural networks \cite{psichogios1994svd} uses a two-stage process for training, \ds{by} \dd{optimizing} the weights of a single layer network  in one stage and the biases  in the other stage, iteratively. In the second stage, \dd{small singular values in a linear least squares problem} indicate redundant neurons \dd{that} can be eliminated. This method only applies to single layer networks and \dd{ignores} redundancies in the weight matrix.

\dd{For} single-hidden-layer neural networks, \cite{teoh2006estimating} studied and related the rank of the weight matrix to the complexity of the decision boundaries of a trained network,
\dd{adding one neuron when there is no distinct gap in the singular values of the weight matrix and removing a number of neurons when there is rank deficiency.} 
\dd{After each change, they train the new network from scratch.}
\ds{Our algorithms are applicable to multilayer networks and allow addition or deletion of multiple neurons at once. In the case of a previously-trained network, we identify specific neurons to prune and keep the rest of the trained network intact.}

SVD is used by \cite{xue2013restructuring} to restructure deep network acoustic models. Their approach discards small singular components of the weight matrices and replaces each layer in the network with two new layers, one purely linear, each with fewer nodes than the original single layer.  This results in a smaller number of parameters if there are many redundant nodes in the network but a larger number of parameters if there is little redundancy. They then use additional training if necessary. \cite{chung2016simplifying} took a similar approach using SVD of weight matrices and replacing each layer in the network with two new layers, but instead of discarding small singular values they sparsify the weight matrices. These approaches do not address the problem of setting the initial structure of the network.  \dd{In Algorithm \ref{alg_3} we use pivoted QR decomposition to reveal candidates} for neurons that can be removed from a trained model in a faster and more effective manner that does not require adding new layers to the network.


\dd{In \cite{alvarez2017compression},  an SVD-based regularization term encourages rank-deficiency in the  weight matrices, identifying layers to compress.}
SVD has also been used in methods that reconstruct a compact version of a trained neural network, \dd{e.g., \cite{denton2014exploiting,goetschalckx2018efficiently,xu2018trained}.} These methods \dd{begin with} an oversized but accurate trained model. As their authors explain, these  \dd{are methods to produce a compact version of a trained network rather than to design the structure of a network}.






\dd{SVD has also been used} in convolutional neural networks with a fixed structure \ds{to} control the behavior of the Jacobian matrix of the function computed by the neural network, enabling better behavior of the optimization algorithms used for training \cite{sedghi2018the}.

\subsection{\dd{A note on cost}}
\label{sec_cost}
Our goal in this work is to reduce the human {and computational} cost of designing deep learning models {in order to facilitate their use in real-world} applications.
Our work is summarized in three algorithms. {The first two algorithms are for designing a neural network from scratch. The first eliminates possibly redundant neurons in order to determine a proper proportion of neurons layer by layer, using partial training of network.} The second  scales a neural network up or down, again with partial training, preserving the proportions determined by the first algorithm, and chooses a size with low validation and generalization errors. The third is applied to a fully-trained network to remove redundant neurons.

\dd{The main tool in our approach is matrix decomposition, in particular, pivoted QR decomposition, rank-revealing QR decomposition, or SVD of the weight-bias matrices for the neural network.
It is important to note that {\em the effort needed for any of these decompositions for dense matrices is negligible compared to the overall training process.} At each step of training for each mini-batch, the derivative of the loss function is computed {for individual training points,} with respect to each and every element in the weight matrices, which involves multiplication of weight matrices. The complexity of computing singular values or QR decomposition is of the same order (if exact algorithms are used) or less (if approximate or early-termination algorithms are used).}

\section{Framework} \label{sec_framework}

\begin{figure}[h]
\includegraphics[width=.7\columnwidth]{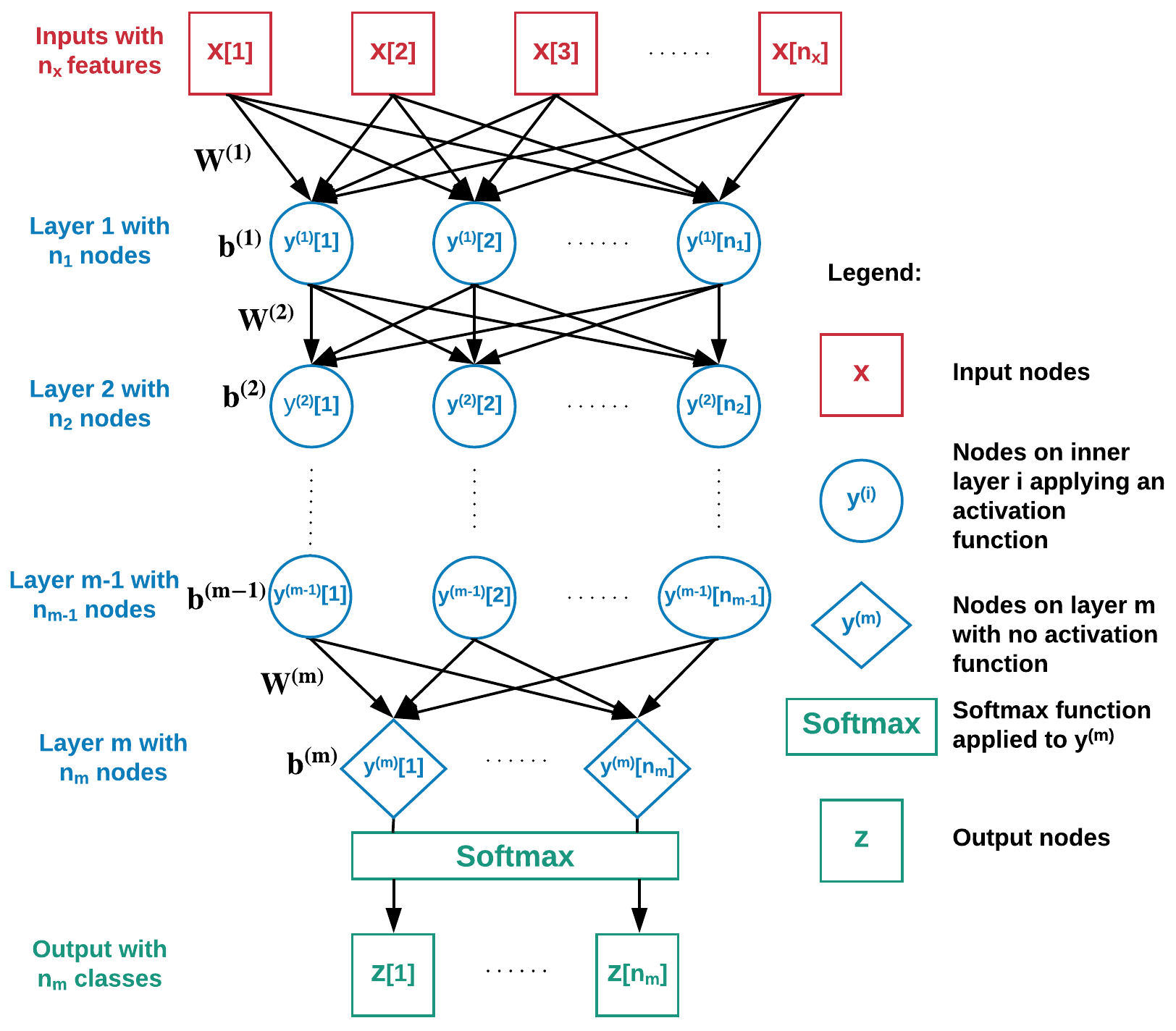}
\centering
\caption{Sketch of a prototype feed-forward neural network $\mathcal{N} $ with $n_x$ inputs, $m$ layers, and $n_m$ outputs.} 
\label{fig_network}
\end{figure}

We explain our method for the neural network prototype $\mathcal{N} $ shown in Figure \ref{fig_network}, as an example. Our method can be easily generalized to neural networks with different architectures, such as convolutional and residual networks. 

In our notation, vectors and scalars are in lower case and matrices are in upper case. Bold characters are used for vectors and matrices, and the relevant layer in the network is shown as a superscript in parenthesis. Subscripts denote the index for a particular element of a matrix or \ds{vector.} 

\dd{We specify a neural network $\mathcal{N}$ by  weight matrices $\bs{W}^{(i)}$ and bias vectors $\bs{b}^{(i)}$ for each layer $i=1,\dots,m$. The input to our neural net is $\bs{y}^{(0)} = \bs{x}$.
 The $i$th layer applies the activation function to the input  $\bs{y}^{(i-1)} \bs{W}^{(i)} + \bs{b}^{(i)}$ to produce output $\bs{y}^{(i)}$.}
 Any of the typical activation functions can be used: sigmoid, relu, erf, etc.
 Inputs are denoted by the vector $\bs{x}$.
We use a training function $\mathcal{T}$, specified by
$$[\mathcal{\hat{N}}, \epsilon^{tr}, \epsilon^{v}] = \mathcal{T}(\mathcal{N},\mathcal{D}^{tr}, \mathcal{D}^{v}, \eta)$$ to train an existing neural network $\mathcal{N}$ using $\eta$ epochs.  Here, $\mathcal{D}^{tr}$ is the training set and  $\mathcal{D}^{v}$ is the validation set. $\mathcal{T}$ returns the trained network $\mathcal{\hat{N}}$, and also the accuracies  $\epsilon^{tr}$  on $\mathcal{D}^{tr}$ and $\epsilon^{v}$  on $\mathcal{D}^{v}$. \olive{Networks that are partially or fully trained are distinguished with $\mathcal{\hat{N}}$ while untrained networks are shown as $\mathcal{N}$.}

\section{\olive{Designing the structure of a neural network by adaptive restructuring}} \label{sec_design}

\olive{We would like to design our network so that it learns a training set and generalizes well, i.e., performs well on a validation or a testing set. Given the framework described above, the main goal is to find the number of neurons needed on each layer of the network, in order for the network to generalize well. Training many possible network structures and choosing the best model can be a very expensive approach as mentioned earlier. Here, we take a more insightful approach based on matrix conditioning of trainable parameters.}

\dd{Let $\bs{\hat{W}}^{(i)}$ denote the \ds{``stacked"} matrix formed by appending the row vector $\bs{b}^{(i)}$ to $\bs{W}^{(i)}$.} \rblu{We use $\kappa(\cdot)$ as a function to compute the 2-norm condition number of matrices.} We build our design method based on two insights:
\begin{itemize}
\olive{
\item If a parameter matrix $\bs{\hat{W}}^{(i)}$ has high condition number compared to other layers, or it is close to rank deficient, this can indicate that the number of neurons on layer $i$ is over-proportioned, compared to other layers. In such cases, we make layer $i$ smaller, so that its share of the overall number of neurons becomes proportionate. We repeat this process until all the matrices have roughly small and similar condition numbers, implying that all layers have the right proportion of neurons.
\item Once we have found the distribution of neurons among the layers of a network, the network might still be over-sized or under-sized, so we scale the size of network up or down, maintaining the same proportion of neurons for the layers. By partial training of such networks, we find the network that performs best on a validation/testing set.
}
\end{itemize}
These two insights \dd{lead to} Algorithm \ref{alg_1} and Algorithm \ref{alg_2}, \dd{which we now present.}

\subsection{Finding a distribution of neurons that leads to small condition numbers among the layers of network}

\dd{To make use of our first insight, we need to compute the numerical rank of $\bs{\hat{W}}^{(i)}$, i.e., the number of sufficiently large singular values. This can be computed using the SVD or estimated using
approximation algorithms, rank-revealing QR decomposition, or pivoted QR decomposition.
\ds{The two QR algorithms compute} an orthogonal matrix $\bs{Q}$, an upper-triangular matrix $\bs{R}$, and a permutation matrix $\bs{P}$ so that 
$$\bs{\hat{W}}\bs{P} = \bs{Q} \bs{R}.$$
Multiplying 
$\bs{\hat{W}}$ by $\bs{P}$ pulls the columns of $\bs{\hat{W}}$ deemed most linearly independent (non-redundant) to the left. The magnitudes of the main diagonal elements of $\bs{R}$ are non-increasing, so we can stop the decomposition when a main diagonal element becomes too small relative to the first.}

%
\dd{In Algorithm \ref{alg_1}, we reduce} the number of neurons on each layer of the network until \dd{all of the matrices have condition number less than $\tau$.} 
\ds{Although written in terms of the SVD, a rank-revealing QR could be used instead.}
If we want the network to have a round number of neurons, we can enforce this condition as we remove neurons. After reducing the number of neurons, we train the new network with $\eta$ epochs.  \dd{In our numerical experiments, \ds{$\eta \le 3$ epochs were} sufficient to identify redundant neurons.} \olive{Note that at this stage, we are not concerned about the accuracy of models and our focus is on the values and variations of condition numbers among the layers.} Algorithm \ref{alg_1} works based on partial training, does not compute the accuracies, and the steps it takes at each iteration does not necessarily improve the accuracy, especially when it is working on an undersized network. It merely adjusts the number of neurons among the layers of the network such that the number of neurons for each layer is well proportioned compared to the others. In the next stage, we take the accuracies into account.

\begin{algorithm}[h]
\caption{Algorithm for \olive{determining distribution of neurons among layers} of a feed-forward neural network}
\label{alg_1}
\textbf{Inputs}: Initial neural network $\mathcal{N}$, $\eta$ , $\tau$ , $\mathcal{D}^{tr}$  \\
\textbf{Output}: Neural network with well-conditioned parameter matrices
\begin{algorithmic}[1] 
\STATE $ \hat{\mathcal{N}} = \mathcal{T}(\mathcal{N},\mathcal{D}^{tr}, [-], \eta )$
\WHILE{ any weight matrix $\bs{\hat{W}}^{(i)}$ of $\hat{\mathcal{N}}$ has condition number $> \tau$}
	\FOR{all such weight matrices}
		\STATE If $\bs{\hat{W}}^{(i)}$ of $\hat{\mathcal{N}}$ has $p$ singular values  less than $ 1/ \tau$ times the largest one, 
	             then remove $p$ neurons from layer $i$ in \ds{${\mathcal{N}}$}.
	\ENDFOR
	\STATE $ \hat{\mathcal{N}} = \mathcal{T}(\ds{{\mathcal{N}}},\mathcal{D}^{tr}, [-], \eta )$
\ENDWHILE 
\STATE \textbf{return} \rblu{${\mathcal{\hat{N}}}$}
\end{algorithmic}
\end{algorithm}


\begin{algorithm}[h]
\caption{Algorithm for optimizing the overall number of neurons in a neural network, while maintaining the proportion of neurons among the layers}
\label{alg_2}
\textbf{Inputs}: Base model $\mathcal{N}^{0}$ (obtained from Algorithm \ref{alg_1}), $\{ \beta_1, \dots , \beta_p \}$ , $\mathcal{D}^{tr}$, $\mathcal{D}^{v}$, $\eta$ , $q$ \\
\textbf{Outputs}: Refined trained $\hat{\mathcal{N}}$
\begin{algorithmic}[1] 
\FOR{$j = 1$ to $p$}
	\STATE Change the number of neurons in all hidden layers of $\mathcal{N}^{0}$, by a factor of $\beta_j$, to obtain $\mathcal{N}^{j}$.
	\FOR {$l = 1$ to $q$}
		\STATE $ [\hat{\mathcal{N}}^{j}, \epsilon_l^{j,tr}, \epsilon_l^{j,v}] = \mathcal{T}(\mathcal{N}^{j},\mathcal{D}^{tr}, \mathcal{D}^{v}, \eta )$
	\ENDFOR
	\STATE $\hat{\epsilon}^{j,tr} = \frac{1}{q} \sum_{l=1}^q \epsilon_l^{j,tr}$
	\STATE $\hat{\epsilon}^{j,v} = \frac{1}{q} \sum_{l=1}^q \epsilon_l^{j,v}$
\ENDFOR
\STATE Choose the model that has the least $2\hat
{\epsilon}^{j,v} - \hat{\epsilon}^{j,tr}$ as $\hat{\mathcal{N}}$.
\STATE Fully train $\hat{\mathcal{N}}$. 
\STATE \textbf{return} $\hat{\mathcal{N}}$
\end{algorithmic}
\end{algorithm}

\subsection{Scaling the size of a neural network}
After we have the right proportion of neurons on each layer, we can expand or contract the neural network, maintaining these proportions. The goal here is to find the overall number of neurons needed to achieve the highest accuracy possible for the model.
We need to estimate the generalization error as we modify the number of neurons. Therefore, we reserve part of the training set as a validation set, if a separate validation set is not available.

In Algorithm \ref{alg_2}, we begin with  a base model $\mathcal{N}^0$, possibly obtained from Algorithm \ref{alg_1}, with a good proportion of neurons on each layer. Given a set of positive scalars $\{ \beta_1, \dots , \beta_p \}$, we construct $p$ new models, where $\mathcal{N}^j$,  increases or decreases the number of neurons in all layers of the base network $\mathcal{N}^0$ by a factor $\beta_j$. This way we obtain $p+1$ models of different size, with the same \dd{relative} distribution of neurons on their layers. Each model is trained $q$ separate times, from scratch, using $\eta$ epochs, and the errors on the training and validation sets are averaged to obtain $\hat{\epsilon}^{j,tr}$ and $\hat{\epsilon}^{j,v}$, where $j \in \{ 1, \dots, p \}$. We use $q=5$ in our computations, since no significant change was observed when using larger values.

Among these $p+1$ models, we  \dd{choose the model that has the least sum} of the validation error, $\hat
{\epsilon}^{j,v}$, and the generalization error, $\hat
{\epsilon}^{j,v} - \hat{\epsilon}^{j,tr}$. 
This procedure is formalized as Algorithm \ref{alg_2}.
By finding the model that minimizes $2\hat{\epsilon}^{j,v} - \hat{\epsilon}^{j,tr}$, we avoid over-fitting and under-fitting in the model. If the smallest or largest model happens to have the smallest $2\hat
{\epsilon}^{j,v} - \hat{\epsilon}^{j,tr}$, we could extend our investigation beyond the $p$ models, by adding more $\beta$'s in the direction of smaller or larger models.

\section{Squeezing trained networks} \label{sec_squeeze}
The two algorithms in the previous section design and train  a network from scratch, given the desired number of layers. \dd{We} now introduce a method to squeeze networks that are already trained but have excess neurons. The method we propose in Algorithm \ref{alg_3} does not necessarily retain the accuracy of the trained model, but it preserves the main essence of it. In our numerical results, we demonstrate that squeezed networks either closely retain the accuracy, or they can be retrained to the best accuracy very quickly.
\dd{As in Algorithm \ref{alg_1}, we need to identify and remove redundancies.
In this case, though, we do not want to discard the result of previous training, so we use the pivoted QR decomposition to tell us which neurons (i.e., which columns of $\hat{W}^{(i)}$) to retain.}
%
%
The parameter $\tau$ defines the threshold for excessive neurons. If \dd{$\tau$} is large, the output of Algorithm \ref{alg_3} can be the same network as the input, and the user might then choose to reduce $\tau$. 

Note that we do not need to compute \dd{the full QR decomposition}; we can stop when a diagonal element of $\bs{R}$ becomes too small.

\begin{algorithm}[h]
\caption{Algorithm for squeezing a trained feed-forward neural network}
\label{alg_3}
\textbf{Inputs}: Trained neural network $\mathcal{\hat{N}}$, $\tau$ , $\mathcal{D}^{tr}$  \\
\textbf{Outputs}: Squeezed neural network with same or smaller number of~neurons
\begin{algorithmic}[1] 
\FOR {$i = 1$ to $m$}
	\IF {\rblu{$\tau < \kappa(\bs{\hat{W}}^{(i)})$ }} \label{alg_3_if}
		\STATE $ [\bs{Q},\bs{R},\bs{P}] = \text{QR} \big(\bs{\hat{W}}^{(i)} \big)$
		   \STATE Define $p$ so that 
		              $|r_{p+1,p+1}| < \tau |r_{11}|$ and
		              $| r_{pp}| \ge \tau |r_{11}|$ \label{alg_3_p}
		   \STATE Remove columns $p+1 : n_i$ of $\bs{P}$ from $\bs{\hat{W}}^{(i)}$
		   \WHILE{$\tau < \rblu{\kappa(\bs{\hat{W}}^{(i)})}$}
		   	\STATE $p = p - 1$
			\STATE Remove column $p+1$ of $\bs{P}$ from $\bs{\hat{W}}^{(i)}$
		   \ENDWHILE
		   \STATE Remove neurons  $p+1 : n_i$ of $\bs{P}$ from the network, by removing corresponding columns of $\bs{{W}}^{(i)}$, rows of $\bs{{W}}^{(i+1)}$, and elements of $\bs{{b}}^{(i)}$
	\ENDIF 
\ENDFOR
\STATE Improve $\mathcal{\hat{N}}$ by retraining, if desired.
\STATE \textbf{return} squeezed $\mathcal{\hat{N}}$
\end{algorithmic}
\end{algorithm}

Our squeezing method is straightforward and simple to use. Unlike methods that rebuild a trained model using specialized training methods, 
\olive{we keep \ds{the trained network intact except for redundancies}. After squeezing, one can retrain the obtained network with a few epochs, which sometimes leads to even better accuracy. For retraining in our approach, one can use the original method of training, and there would be no necessity for specific loss functions and optimization methods.}

\dd{It is important to remember, as mentioned in section \ref{sec_cost}, that the} effort needed for computation of \dd{pivoted QR decomposition of}  the weight matrices of a network is negligible compared to the overall training process. 
%
We recommend our Algorithm \ref{alg_3} as a computationally inexpensive and approachable method to squeeze trained networks and to gain insight about their compressibility. Other sophisticated methods that rebuild the networks from scratch may have certain advantages in particular applications, but their computational cost \dd{may be} much higher.

In Algorithm~\ref{alg_3}, use of pivoted QR or RR-QR is necessary because we need to know which specific neurons are redundant, the information we obtain from the permutation matrix of decomposition. The while loop in our algorithm makes sure the condition number of resulting matrices are below the~$\tau$, after elimination of redundant neurons. This is because line \ref{alg_3_p} of our algorithm may possibly overestimate the rank of matrix leading to elimination fewer than necessary neurons. This while loop usually takes zero or very few iterations, \rblu{and it is not an essential part of the algorithm. In cases where a network is squeezed, retrained, and squeezed again, using the while loop may reduce the overall cost, because it could cause line \ref{alg_3_if} of the algorithm not to be invoked in the subsequent squeeze. In other cases, where squeezing is applied once, the while loop can be dropped.}

\section{Numerical results} \label{sec_results}

In our numerical results, we use TensorFlow to train the networks, with Adam optimizer and learning rate of 0.001. We also use a tunable error function as the activation function, but keep in mind that our training method does not depend on the choice of activation function. We start with MNIST which can be considered an unfamiliar dataset, because we use the wavelet coefficients of images, instead of the pixel data.
		
\subsection{MNIST} \label{sec_mnist}

The MNIST dataset has 10 output classes, corresponding to the digits 0 through 9. We represent each data point as a vector of length 200, using the Haar wavelet basis. The 200 most significant wavelets are chosen by rank-revealing QR decomposition of the matrix formed from the wavelet coefficients of all images in the training set. Using this small number of wavelet coefficients and a simple feed-forward network will lead to accuracy of about 98.7\%. Accuracy could be improved using more wavelet coefficients, and using regularization techniques in the literature, but this accuracy is adequate to demonstrate the effectiveness of our method.

\textbf{Using Algorithm \ref{alg_1}.} 
We consider a neural network of 12 hidden layers with 300 nodes on each layer as the input to Algorithm \ref{alg_1}. After the initial training of this model, the condition numbers of the stacked weight matrices vary between $2$ and $2,652$, as shown in the third column in Table \ref{tbl_1}.  
We use Algorithm \ref{alg_1} to adjust the proportions of neurons, with $\tau = 25$ and $\eta = 1$. The number of neurons and the condition numbers of the matrices for the output of Algorithm \ref{alg_1} are presented in the last two columns in Table \ref{tbl_1}. At each iteration, we have rounded down the number of neurons obtained at line 4 of the algorithm to a multiple of 5.

\renewcommand{\baselinestretch}{1.2}
\setlength{\tabcolsep}{5pt}
\begin{table}[h]
\centering
\small
\caption{Condition numbers of the stacked matrices and the number of neurons on each of the 12 layers of the network  processed by Algorithm \ref{alg_1} to learn 200 wavelet coefficients for MNIST. }
\begin{tabular}{c|rr|rr}  
\toprule
\multirow{2}{*}{\bf Layer $(i)$} & \multicolumn{2}{c|}{\bf Initial network }  &  \multicolumn{2}{c}{\bf Algorithm \ref{alg_1}} \\ 
 & $n_i$ & $\kappa(\bs{\hat{W}}^{(i)})$ & $n_i$ & $\kappa(\bs{\hat{W}}^{(i)})$ \\ \hline
1 & 300 & 10 & 300 & 9  \\
2 & 300 & 649 & 205 & 10  \\
3 & 300 & 301 & 255 & 15  \\
4 & 300 & 2,652 & 210 & 20 \\
5 & 300 & 275 & 250 & 22  \\
6 & 300 & 583 & 210 & 23  \\
7 & 300 & 946 & 180 & 24  \\
8 & 300 & 268 & 150 & 24  \\
9 & 300 & 433 & 120 & 17  \\
10 & 300 & 1,269 & 95 & 14  \\
11 & 300 & 398 & 65 & 11  \\
12 & 300 & 673 & 25 & 4 \\
13 & 10 & 2 & 10 & 4  \\
\bottomrule
\end{tabular}
\label{tbl_1}
\end{table}
\renewcommand{\baselinestretch}{2}

We observe that the final condition numbers are relatively close to each other and less than $\tau$. Additionally, they monotonically increase towards the middle layer and then monotonically decrease towards the output layer. This monotonicity of condition numbers is not a requirement and might not be achieved for all models.

\textbf{Using Algorithm \ref{alg_2}.} The previous step found a promising set of proportions for the sizes of the layers. Using the output of Algorithm \ref{alg_1}, given in  Table \ref{tbl_1}, as our base model, we scale this network to try to improve the accuracy. We chose eight $\beta$'s ranging from $1$ to $2.4$, with increments of $0.2$.

For this step, we need a validation set. Hence, we remove 10,000 images from the training set, randomly selecting 1,000 images from each class to use as a validation set $\mathcal{D}^{v}$. This leaves the training set $\mathcal{D}^{tr}$ with only 50,000 images.
  
\begin{figure}[h]
\includegraphics[width=.55\columnwidth]{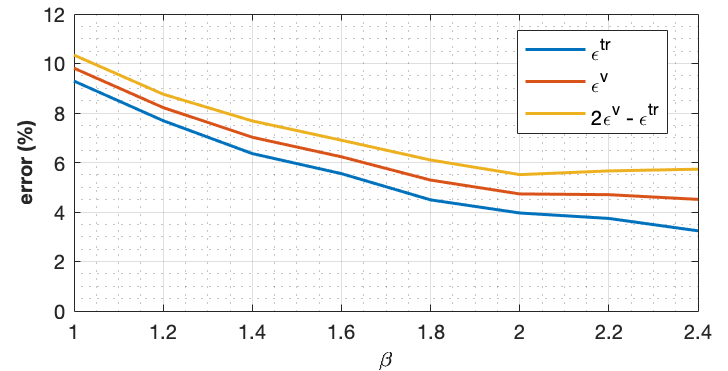}
\centering
\vspace{-.3cm}
\caption{Errors of the eight models investigated by Algorithm \ref{alg_2} for MNIST. We have chosen the model with $\beta = 2.0$, because it has  the least sum of validation and generalization errors, when models are partially trained with 1 epoch.} 
\label{fig_mnist_gen}
\end{figure}

We use a batch size of 50, and set $q = 5,$ $\eta = 1$. Algorithm \ref{alg_2} partially trains all eight models to achieve the errors shown in Table \ref{tbl_2} and Figure \ref{fig_mnist_gen}. It then chooses the model with $\beta = 2$ as the best model and trains it using all 60,000 images in the training set, to achieve 100\% and 98.68\% accuracies on the training and testing sets, respectively. Achieving this accuracy with such a small neural network is remarkable, considering that we only used 200 wavelet coefficients, and we did not use any regularization or any sophisticated architecture for the network.

\renewcommand{\baselinestretch}{1.2}
\setlength{\tabcolsep}{7pt}
\begin{table}[h]
\centering
\small
\caption{Errors of the eight networks obtained from Algorithm \ref{alg_2}, defined by the $\beta$'s, partially trained on the reduced training set (with 50,000 images) and validated using 10,000 images.}
\begin{tabular}{cccr}  
\toprule
$\bs{\beta}$ & $\bs{\hat{\epsilon}^{tr}}$ & $\bs{\hat{\epsilon}^{v}}$ & $\bs{2\hat{\epsilon}^{v} - \hat{\epsilon}^{tr}}$  \\
\midrule
1.0 & 9.29 & 9.81 & 10.34  \\
1.2 & 7.69 & 8.22 & 8.76  \\
1.4 & 6.36 & 7.02 & 7.68  \\
1.6 & 5.55 & 6.23 & 6.90  \\
1.8 & 4.49 & 5.29 & 6.10  \\
2.0 & 3.96 & 4.73 & {\bf 5.51}  \\
2.2 & 3.74 & 4.70 & 5.66  \\
2.4 & 3.24 & 4.51 & 5.73  \\
\bottomrule
\end{tabular}
\label{tbl_2}
\end{table}

\renewcommand{\baselinestretch}{2}

For all of these models, the condition numbers of the stacked matrices are similar to those presented in the 5th column of Table \ref{tbl_1} and smaller than 25. This indicates that partial training with $\eta = 1$ has adequately captured the \dd{conditioning} of the matrices. It also indicates that scaling the number of neurons, while maintaining their proportions layer-by-layer,  has little effect on the condition numbers for the stacked matrices.

\textbf{Verifying the results.} Here, we investigate whether the model we obtained with $\beta = 2.0$ is in fact the best network we can choose from the pool of networks defined by the eight $\beta$'s. For this, we fully trained all eight of the  networks, obtaining the errors in Table \ref{tbl_3}. Evidently, the best accuracy is achieved by the model chosen by Algorithm \ref{alg_2}, confirming the effectiveness of our method.

\renewcommand{\baselinestretch}{1.2}
\setlength{\tabcolsep}{6pt}
\begin{table}[h]
\centering
\small
\caption{Testing error when models are fully trained with all 60,000 images in the MNIST training set}
\begin{tabular}{ccccccccc}  
\toprule
$\bs{\beta}$ & 1 & 1.2 & 1.4 & 1.6 & 1.8 & 2.0 & 2.2 & 2.4  \\
\midrule
$\bs{\hat{\epsilon}^{te}}$ &  1.97 &  1.76 &  1.65 &  1.56 &  1.43 &  {\bf 1.32} & 1.56 &  {1.61}  \\
\bottomrule
\end{tabular}
\label{tbl_3}
\end{table}
\renewcommand{\baselinestretch}{2}

Clearly, trying to squeeze this trained network using Algorithm \ref{alg_3} with $\tau \geq 25$ will have no effect. Overall, we observed  that choosing a value of $\tau$  between 20 and 50 in Algorithm \ref{alg_1}, and then applying Algorithm \ref{alg_2}, leads to similar networks with similar best accuracies. However, choosing $\tau$ outside of this range leads to models with slightly inferior accuracies. The key factor in choosing a good value for $\tau$ \dd{seems to be} the variance of condition numbers among the layers. The values of $\tau$ that deliver the best results also yield condition numbers with small variance among the layers. 
This approach would yield similar results as when we choose the value $\tau$ between $30$ and $40$ in the first place. This range can be viewed as a practical choice for~$\tau$.

\subsection{Adult Income dataset} \label{sec_adult}
Next, we consider the Adult Income dataset from the UCI Machine Learning Repository \cite{Dua2017}, an example that has a combination of discrete and continuous variables. There are 32,561 data points in the training set and 16,281 in the testing set. Each data point has information about an individual, and the label is binary, indicating whether the individual's income is greater than \$50K annually.

Each of the continuous variables (age, fnlwgt, education-num, capital-gain, capital-loss and hours-per-week) has a lower bound of 0.  We normalize each variable to the range 0 -- 100 using upper bounds of 100, 2e6, 25, 2e5, 1e4 and 120, respectively. Moreover, we transform the categorical variables (workclass, education level, marital status, occupation, relationship, race, sex, native country) into a binary form where each category type is represented by one binary feature. The categories that are active for a data point have binary value of 1 in their corresponding features, while the \dd{others} are set to zero.

\textbf{Using Algorithm \ref{alg_1}.} 
We consider a neural network of 12 hidden layers with 50 nodes on each layer as the input to Algorithm \ref{alg_1}. Similar to the previous section, the properties of the initial and final network  are presented in Table \ref{tbl_4}. 
For this dataset, we have not rounded the number of neurons, and we have used batch size of 20, $\eta = 3$, and $\tau = 40$. \olive{This time, we choose a larger $\eta$ compared to previous example, because our network is much smaller and training with the larger $\eta$ still takes just a few seconds. We also choose a larger $\tau$ because the condition numbers tend to remain large during the process. We discuss the factors involved in choosing these parameters further in \Cref{sec_hyper}.}

\renewcommand{\baselinestretch}{1.2}
\setlength{\tabcolsep}{5pt}
\begin{table}[h]
\centering
\small
\caption{Condition numbers of the stacked matrices and the \ds{number} of neurons on each of the 12 layers of the network  processed by Algorithm \ref{alg_1} to learn the Adult Income dataset. }
\begin{tabular}{c|rr|rr}  
\toprule
\multirow{2}{*}{\bf Layer $(i)$} & \multicolumn{2}{c|}{\bf Initial network }  &  \multicolumn{2}{c}{\bf Algorithm \ref{alg_1} } \\ 
 & $n_i$ & $\kappa(\bs{\hat{W}}^{(i)})$ & $n_i$ & $\kappa(\bs{\hat{W}}^{(i)})$ \\ \hline
1 & 50 & 9 & 44 & 7  \\
2 & 50 & 654 & 39 & 37  \\
3 & 50 & 658 & 32 & 31  \\
4 & 50 & 583 & 22 & 13 \\
5 & 50 & 230 & 20 & 30  \\
6 & 50 & 224 & 15 & 13  \\
7 & 50 & 159 & 12 & 20 \\
8 & 50 & 912 & 8 & 20  \\
9 & 50 & 136 & 5 & 14 \\
10 & 50 & 377 & 4 & 7  \\
11 & 50 & 74 & 8 & 16  \\
12 & 50 & 110 & 6 & 18 \\
13 & 2 & 1 & 2 & 3  \\
\bottomrule
\end{tabular}
\label{tbl_4}
\end{table}
\renewcommand{\baselinestretch}{2}

\textbf{Using Algorithm \ref{alg_2}.} Using the proportions found in the previous step, we consider eight $\beta$'s ranging from $0.6$ to $2.0$, with increments of $0.2$. For the validation set $\mathcal{D}^{v}$, we randomly remove 10\% of the data points from the training set, leaving the training set $\mathcal{D}^{tr}$ with 90\% of its data points.

\begin{figure}[h]
\includegraphics[width=.55\columnwidth]{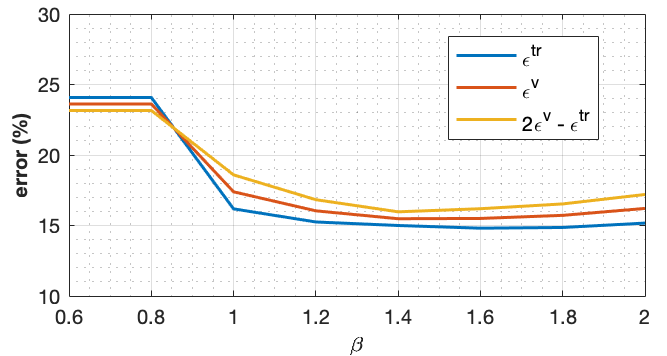}
\centering
\vspace{-.3cm}
\caption{Errors of the eight models investigated by Algorithm \ref{alg_2} for the Adult Income dataset. We have chosen the model with $\beta = 1.4$, because it has the least sum of validation and generalization errors when trained partially.} 
\label{fig_adult_gen}
\end{figure}

Based on the results of Algorithm \ref{alg_2}, shown in Figure \ref{fig_adult_gen}, we choose the neural network with $\beta = 1.4$. After fully training this model we achieve 86.05\% accuracy on the testing set, which is comparable to the best accuracies reported in the literature \cite{friedler2019comparative,mothilal2019explaining}.

\renewcommand{\baselinestretch}{1.2}
\renewcommand{\baselinestretch}{2}

\textbf{Verifying the results.} To verify the results, we fully train all the models defined by the eight $\beta$'s to achieve the testing errors in Table \ref{tbl_adlt_testr}. We observe that the best accuracy is indeed achieved by the model with $\beta = 1.4$. 

\renewcommand{\baselinestretch}{1.2}
\setlength{\tabcolsep}{7pt}
\begin{table}[H]
\centering
\small
\caption{Testing error when models are fully trained with all data in the Adult Income training set }
\begin{tabular}{ccccccccc}  
\toprule
$\bs{\beta}$ & 0.6 & 0.8 & 1.0 & 1.2 & 1.4 & 1.6 & 1.8 & 2.0  \\
\midrule
$\bs{\hat{\epsilon}^{te}}$ &  14.34 &  14.19 &  14.22 &  14.16 &  {\bf 13.95} &  14.14 & 14.19 & 14.32  \\
\bottomrule
\end{tabular}
\label{tbl_adlt_testr}
\end{table}
\renewcommand{\baselinestretch}{1.2}


\subsection{Using Algorithm \ref{alg_3} to squeeze networks trained on MNIST}

\subsubsection*{Squeezing the model obtained via Algorithms \ref{alg_1} and \ref{alg_2} may lead to even better accuracy} 
We first consider the refined network we trained with $\beta = 2$ that achieved 98.68\% accuracy on the testing set. For this model, all the condition numbers $\kappa(\bs{\hat{W}}^{(i)})$ happen to be less than 23. Hence, we squeeze the model with $\tau$ ranging between 22 and 18. Table \ref{tbl_sq_mnist1} shows the results. After squeezing, we measure the accuracy of the model on the testing set and then retrain it, stopping when we reach 100\% accuracy on the training set.

\renewcommand{\baselinestretch}{1.2}
\setlength{\tabcolsep}{3pt}
\begin{table}[h]
\caption{Number of neurons and accuracies of the model with $\beta = 2$, trained in Section \ref{sec_mnist}, squeezed by Algorithm \ref{alg_3} with different values of $\tau$.} 
\centering
\small
\begin{tabular}{cccc}  
\toprule
\parbox[c]{1cm}{\centering $\bs{\tau}$} & \parbox[c]{2.5cm}{\centering \bf Number of neurons removed} & \parbox[c]{4cm}{\centering \bf Accuracy of\\ squeezed model before\\ retraining} & \parbox[c]{4cm}{\centering \bf Accuracy of\\ squeezed model after\\ retraining}  \\
\midrule
22 & 5 & 98.47 & 98.68 \\
20 & 408 & 90.35 & {\bf 98.74} \\
18 & 502 & 80.94 & 98.70 \\
\bottomrule
\end{tabular}
\label{tbl_sq_mnist1}
\end{table}
\renewcommand{\baselinestretch}{2}

It is notable that retraining the squeezed models may lead to accuracies better than the original model. Table \ref{tbl_sq_mnist_best} shows the size and conditioning of  the best model, with accuracy of 98.74\%.

\renewcommand{\baselinestretch}{1.2}
\setlength{\tabcolsep}{5.2pt}
\begin{table}[h]
\centering
\small
\caption{Condition numbers of the stacked matrices and the number of neurons on each layer of the model with 98.74\% accuracy on MNIST. }
\begin{tabular}{l rrrrrrrrrrrrr}  
\toprule
{\bf Layer $(i)$} & 1 & 2 & 3 & 4 & 5 & 6 & 7 & 8 & 9 & 10 & 11 & 12 & 13 \\
\midrule
$\bs{n}_i$ & 600 & 410 & 510 & 420 & 364 & 319 & 278 & 249 & 212 & 180 & 130 & 50 & 10 \\
$\bs{\kappa}(\bs{\hat{W}}^{(i)})$ & 4 & 10 & 17 & 18 & 20 & 20 & 20 & 20 & 19 & 19 & 10 & 4 & 2   \\
\bottomrule
\end{tabular}
\label{tbl_sq_mnist_best}
\end{table}
\renewcommand{\baselinestretch}{2}

\subsubsection*{Squeezing can accurately detect excess neurons} 
Here, we consider the model in Table \ref{tbl_sq_mnist_best}, add 20 neurons on its 4th and 8th hidden layers, and fully train it to achieve 98.47\% accuracy. This decrease in the accuracy can be associated with overfitting, caused by addition of those 40 neurons. Table \ref{tbl_sq_mnist_add20} shows the condition numbers of the stacked matrices for this model. We observe that condition numbers have increased not only for layers 4 and 8, but also for the in-between layers 6 and 7. 

\renewcommand{\baselinestretch}{1.2}
\setlength{\tabcolsep}{5.2pt}
\begin{table}[h]
\centering
\small
\caption{Condition numbers of the stacked matrices for the model that has 20 more neurons on its 4th and 8th layers compared to the model in Table \ref{tbl_sq_mnist_best}. The condition numbers of layers 4,6,7 and 8 have noticeably increased above the $\tau=20$ we had used to squeeze that model.}
 \begin{tabular}{l rrrrrrrrrrrrr}  
\toprule
{\bf Layer $(i)$} & 1 & 2 & 3 & 4 & 5 & 6 & 7 & 8 & 9 & 10 & 11 & 12 & 13 \\
\midrule
$\bs{n}_i$ & 600 & 410 & 510 & 440 & 364 & 319 & 278 & 269 & 212 & 180 & 130 & 50 & 10 \\
$\bs{\kappa}(\bs{\hat{W}}^{(i)})$ & 4 & 10  & 16  & 26  & 19  & 30  & 30  & 75  & 16  & 21  & 12  & 4 & 2   \\
\bottomrule
\end{tabular}
\label{tbl_sq_mnist_add20}
\end{table}
\renewcommand{\baselinestretch}{2}

Algorithm \ref{alg_3} enables us to extract some extra neurons from this model, but we have to choose the $\tau$ wisely. By looking at the condition numbers in Table \ref{tbl_sq_mnist_add20}, we see that only layer 8 has condition number $>30$, hence, we squeeze the model with $\tau=30$. The algorithm discards 20 neurons from the 8th layer, leaving a model with 98.45\% accuracy. Retraining this model leads to 98.55\% accuracy.

Similarly, if we squeeze the model with $\tau=25$, a total of 37 neurons will be discarded from the network: 3 neurons from the 4th layer, 4 neurons from the 6th layer, 5 neurons from the 7th layer, and 25 neurons from the 8th layer. The accuracy of the squeezed model is 98.39\% before retraining, and 98.66\% after retraining. 

So, Algorithm \ref{alg_3} enabled us to effectively extract extra neurons from the model and obtain better accuracies.

\subsubsection*{Squeezing an oversized model reduces the overfitting and improves the accuracy} 
As the last experiment on MNIST, we study an oversized network with 600 neurons on each layer. Training this oversized network leads to accuracy of 98.3\% on the testing set, clearly because of overfitting. This model has 3,438 more neurons compared to the model with best accuracy in Table \ref{tbl_sq_mnist_best}, leading to an increase of 199\% in the number of training parameters.

We squeeze this oversized model using Algorithm \ref{alg_3} with different values of $\tau$. The results of squeezing are presented in Table \ref{tbl_sq_mnist_over}. The accuracies of models decrease after squeezing, although after retraining the squeezed models we obtain accuracies as good, or even better than the original oversized model. This improvement demonstrates the effectiveness of Algorithm \ref{alg_3} in reducing the over-fitting and discarding the excess neurons.

\renewcommand{\baselinestretch}{1.2}
\setlength{\tabcolsep}{2.5pt}
\begin{table}[H]
\centering
\small
\caption{Number of neurons removed and the resulting accuracies, after squeezing an oversized model with 600 neurons per hidden layer, using Algorithm \ref{alg_3} with different values of $\tau$.}
\begin{tabular}{cccc}  
\toprule
\parbox[c]{1cm}{\centering $\bs{\tau}$} & \parbox[c]{2.5cm}{\centering \bf Number of neurons removed} & \parbox[c]{4cm}{\centering \bf Accuracy of\\ squeezed model before\\ retraining} & \parbox[c]{4cm}{\centering \bf Accuracy of\\ squeezed model after\\ retraining}  \\
\midrule
500 & 27 & 98.16 & 98.30 \\
200 & 87 & 97.93 & 98.30 \\
100 & 178 & 96.90 &  98.30 \\
50 & 371 & 90.64 & 98.32 \\
40 & 499 & 90.44 &  98.35 \\
35 & 573 & 90.24 & 98.39 \\
30 & 660 & 88.88 & 98.53 \\
25 & 787 & 86.38 & 98.41 \\
20 & 998 & 63.37 & 98.39 \\
\bottomrule
\end{tabular}
\label{tbl_sq_mnist_over}
\end{table}
\renewcommand{\baselinestretch}{2}

After retraining the squeezed models in Table \ref{tbl_sq_mnist_over}, the condition numbers of most matrices go above the $\tau$ used for squeezing, indicating that models are still highly oversized and can be squeezed further to achieve better accuracy. However, this process of squeeze/retrain iterations is less effective than using Algorithms \ref{alg_1} and \ref{alg_2}. So, as a general practice we do not recommend starting the training process with an oversized model.


In the next section, we will further investigate the squeezing process on the Adult Income dataset, and will also study a highly oversized model.

\subsection{Using Algorithm \ref{alg_3} to squeeze networks trained on the Adult Income dataset}

\subsubsection*{Squeezing the model obtained via Algorithms \ref{alg_1} and \ref{alg_2} may lead to even better accuracy} 
Let's consider the best model obtained in Section \ref{sec_adult} with $\beta = 1.4$.
Using Algorithm \ref{alg_3}, we squeeze that model with different values of $\tau$. Clearly, squeezing with $\tau > 40$ will return the exact same model.  Table \ref{tbl_sq_adult} shows the number of neurons removed from the model for three values of $\tau$, along with the accuracy of the squeezed models, before and after retraining.

\renewcommand{\baselinestretch}{1.2}
\setlength{\tabcolsep}{2pt}
\begin{table}[h]
\centering
\small
\caption{Number of neurons and accuracies of the model with $\beta = 1.4$ trained in Section \ref{sec_adult}, after being squeezed by Algorithm \ref{alg_3} with different values of $\tau$. Accuracies of squeezed models have not dropped drastically, and retraining has led to a better accuracy for $\tau = 35$ and $30$.}
\begin{tabular}{cccc}  
\toprule
\parbox[c]{1cm}{\centering $\bs{\tau}$} & \parbox[c]{2.5cm}{\centering \bf Number of neurons removed} & \parbox[c]{4cm}{\centering \bf Accuracy of\\ squeezed model before\\ retraining} & \parbox[c]{4cm}{\centering \bf Accuracy of\\ squeezed model after\\ retraining}  \\
\midrule
35 & 20 & 84.76 & {\bf 86.11} \\
30 & 46 & 78.23 & 86.09 \\
25 & 77 & 76.38 & 86.03 \\
\bottomrule
\end{tabular}
\label{tbl_sq_adult}
\end{table}
\renewcommand{\baselinestretch}{2}

We see that squeezing the model with $\tau = 35$ has led to a model with even better accuracy: 86.11\% on the testing set. 
Properties of this model are presented in Table \ref{tbl_sq_adult_86}. For retraining, we have only used 5 epochs.

\renewcommand{\baselinestretch}{1.2}
\setlength{\tabcolsep}{6pt}
\begin{table}[h]
\centering
\small
\caption{Condition numbers of the stacked matrices and the number of neurons for the model with 86.11\% accuracy on the Adult Income testing set.}
\begin{tabular}{l rrrrrrrrrrrrr}  
\toprule
{\bf Layer $(i)$} & 1 & 2 & 3 & 4 & 5 & 6 & 7 & 8 & 9 & 10 & 11 & 12 & 13 \\
\midrule
$\bs{n}_i$ & 53 & 48 & 40 & 30 & 26 & 21 & 16 & 11 & 7 & 5 & 11 & 8 & 2 \\
$\bs{\kappa}(\bs{\hat{W}}^{(i)})$ & 35 & 19 & 27 & 24 & 33 & 20 & 30 & 12 & 12 & 14 & 14 & 24 & 2  \\
\bottomrule
\end{tabular}
\label{tbl_sq_adult_86}
\end{table}

\subsubsection*{Squeezing an oversized model reduces the overfitting and improves the accuracy} 

Let's consider a large model with 100 neurons on each layer, which has 110,581 more trainable parameters, compared to the model in Table \ref{tbl_sq_adult_86}, an increase of more than 10 times in the number of training parameters. After training this network, we obtain 85.28\% accuracy on the testing set. When using Dropout \cite{srivastava2014dropout} {during the training}, the common approach to avoid overfitting, we achieve 85.45\% accuracy, which is still far less than 86.11\% obtained using our methods.

{Let's now} use Algorithm \ref{alg_3} to squeeze the trained (oversized) model above {with} 85.28\% accuracy, and then retrain it. We perform this squeezing and retraining, with different values of $\tau$, and the corresponding results are presented in Table \ref{tbl_sq_adult2}. Each squeezed model is retrained with 10 epochs. We observe that squeezing with $\tau$ between 30 and 40 has led to best improvements in the accuracy. {This improved accuracy (as a result of squeezing and retraining) is smaller than the best accuracy of 86.11\%, but, it is better than the accuracy of the model trained using Dropout.} We also note that although squeezing makes the condition number of $\bs{\hat{W}}$ matrices $\leq \tau$, the condition numbers can increase above $\tau$ during retraining. This is to be expected, because the squeezed models are still highly oversized.

\renewcommand{\baselinestretch}{1.2}
\setlength{\tabcolsep}{2pt}
\begin{table}[h]
\centering
\small
\caption{Number of removed neurons and accuracies of an oversized model with 100 neurons per hidden layer, after being squeezed by Algorithm \ref{alg_3} with different values of $\tau$.}
\begin{tabular}{cccc}  
\toprule
\parbox[c]{1cm}{\centering $\bs{\tau}$} & \parbox[c]{2.5cm}{\centering \bf Number of neurons removed} & \parbox[c]{4cm}{\centering \bf Accuracy of\\ squeezed model before\\ retraining} & \parbox[c]{4cm}{\centering \bf Accuracy of\\ squeezed model after\\ retraining}  \\
\midrule
100 & 29 & 82.65 & 85.25 \\
50 & 86 & 82.09 & 85.54 \\
40 & 95 & 80.73 &  85.63 \\
35 & 104 & 80.92 & {\bf 85.69} \\
30 & 120 & 79.76 &  85.62 \\
20 & 166 & 76.99 & 85.45 \\
10 & 315 & 76.38 & 85.57 \\
\bottomrule
\end{tabular}
\label{tbl_sq_adult2}
\end{table}
\vspace{0.2in}
\renewcommand{\baselinestretch}{2}

\vspace{-.4cm}
To provide the last insight, let's look into the model squeezed with $\tau = 35$. After retraining, several of its $\bs{\hat{W}}$ matrices have  condition number greater than $35$. We repeat the process, squeezing it with the same $\tau$, and then retraining it with 10 epochs. After 4 squeeze/retrain iterations, we obtain a model that can no longer be squeezed. Table \ref{tbl_sq_adult3} shows how the network has evolved through this  process.

\renewcommand{\baselinestretch}{1.2}
\setlength{\tabcolsep}{5pt}
\begin{table}[h]
\centering
\small
\caption{Number of neurons for a 12-layer network trained on the Adult Income dataset, squeezed using Algorithm \ref{alg_3} with $\tau = 35$ and retrained with 10 epochs, repeatedly, until it cannot be squeezed further. Reported accuracy is on the testing set, after the retraining. Squeezing is computationally inexpensive and significantly improves the accuracy.}
\begin{tabular}{crrrrr}  
\toprule
\multirow{2}{*}{\bf Layer $(i)$} & \multicolumn{5}{c}{\bf Squeeze iteration }  \\ 
\cmidrule(lr){2-6}
& 0 & 1 & 2 & 3 & 4 \\ 
\midrule
1 & 100 & 75 & 72 & 68 & 67  \\
2 & 100 & 100 & 100 & 100 & 100  \\
3 & 100 & 89 & 86 & 86 & 86  \\
4 & 100 & 84 & 80 & 80 & 78 \\
5 & 100 & 100 & 100 & 100 & 100  \\
6 & 100 & 93 & 92 & 92 & 92  \\
7 & 100 & 87 & 84 & 84 & 84 \\
8 & 100 & 100 & 100 & 100 & 100  \\
9 & 100 & 91 & 91 & 90 & 90 \\
10 & 100 & 85 & 85 & 85 & 85  \\
11 & 100 & 100 & 100 & 100 & 100  \\
12 & 100 & 92 & 92 & 92 & 92 \\
\midrule
Accuracy (\%) & 85.3 & 85.68 & 85.70 & 85.73 & 85.81 \\
\bottomrule
\end{tabular}
\label{tbl_sq_adult3}
\end{table}
\renewcommand{\baselinestretch}{2}

The squeezed model has 823 more neurons compared to the model that achieved 86.11\% accuracy. It is also less accurate because of overfitting. However, we should note that this squeezing process improved the accuracy of oversized model from 85.28\% to 85.81\%, which is significant for such a computationally inexpensive process.
This demonstrates the effectiveness of Algorithm \ref{alg_3} in squeezing networks with excessive neurons.

\subsection{Evolution of networks during training and choosing the hyperparameters} \label{sec_hyper}

Here, we provide more information about the evolution of network parameters during the training and provide guidance to choose the hyperparameters in our algorithms.

\subsubsection*{Choosing $\eta$} This parameter is the number of training epochs before refining the network. In Algorithm \ref{alg_1}, if the condition numbers of stacked weight matrices remain mostly similar after a certain number of epochs, then it would be inefficient to choose an $\eta$ larger than that number of epochs. In our experiments, even half of an epoch captures the condition number closely. Of course, we cannot guarantee that one (or half) epoch will be adequate for all datasets. 
Hence, the user of our algorithm should perform \dd{an initial experiment} to see how the condition numbers of weight matrices change \dd{at each epoch} and then choose a good value for $\eta$ accordingly. If computational resources are abundant, choosing a larger $\eta$ would be a safe approach. Nevertheless, if an insufficient number is chosen for $\eta$, some of the condition numbers might go up again and become disproportionate after the final training of the network. This would prompt the user to either squeeze the obtained network or \dd{repeat} Algorithm \ref{alg_1}  \ds{with} a larger $\eta$.

Tables \ref{tbl_eta_mnist1} and \ref{tbl_eta_mnist2} show the evolution of condition numbers for two different models trained on the MNIST example, along with the corresponding number of neurons that should be removed using $\tau = 30$. The model in Table \ref{tbl_eta_mnist1} is an oversized network, and the model in Table \ref{tbl_eta_mnist2} is undersized, compared to the model we obtained, earlier, with the best accuracy. We can see that condition numbers largely remain the same as we train the models with more epochs.

\renewcommand{\baselinestretch}{1.2}
\setlength{\tabcolsep}{6pt}
\begin{table}[h]
\small
\centering
\caption{Evolution of condition numbers of stacked weight matrices of a 9-layer neural network with 600 neurons per hidden layer, trained on our MNIST example. Network is oversized and we expect hidden layers 2 through 9 to lose neurons. The high condition number of layers compared to the first layer is aligned with our expectation. Notice that this is noticeable even after training with small number of epochs and the number of neurons removed from each layer, $p_i$, does not vary much with respect to $\eta$. \rblu{$\big(\kappa_i = \kappa(\bs{\hat{W}}^{(i)})\big)$}}
\begin{tabular}{crrrrrrrrrrr}  
\toprule
\multirow{3}{*}{\bf Layer $(i)$} & \multirow{3}{*}{$n_i$} & \multicolumn{10}{c}{ $\eta$ }  \\ 
\cmidrule(lr){3-12}
& & \multicolumn{2}{c}{.5} & \multicolumn{2}{c}{1} & \multicolumn{2}{c}{2} & \multicolumn{2}{c}{5} & \multicolumn{2}{c}{10} \\ 
\cmidrule(lr){3-12}
& & $\kappa_i$ & $p_i$ & $\kappa_i$ & $p_i$ & $\kappa_i$ & $p_i$ & $\kappa_i$ & $p_i$ & $\kappa_i$ & $p_i$ \\ 
\midrule
1 & 600 & 4 & 0 & 4 & 0 & 4 & 0 & 4 & 0 & 4 & 0  \\
2 & 600 & 1,050 & 25 & 1,041 & 24 & 1,207 & 24 & 627 & 24 & 843 & 24  \\
3 & 600 & 3,609 & 24 & 1,551 & 24 & 3,398 & 24 & 6,878 & 24 & 1,265 & 24  \\
4 & 600 & 663 & 24 & 642 & 25 & 1,083 & 25 & 5,657 & 25 & 1,317 & 25  \\
5 & 600 & 961 & 26 & 1,664 & 26 & 2,472 & 26 & 661 & 26 & 963 & 26  \\
6 & 600 & 2,212 & 25 & 1,746 & 24 & 1,268 & 24 & 751 & 24 & 1,485 & 24  \\
7 & 600 & 1,206 & 26 & 882 & 25 & 904 & 25 & 1,961 & 25 & 1,667 & 25  \\
8 & 600 & 881 & 24 & 899 & 24 & 802 & 24 & 820 & 24 & 768 & 24  \\
9 & 600 & 537 & 25 & 576 & 26 & 744 & 26 & 1,077 & 26 & 1,948 & 26  \\
10 & 10 & 1 & 0 & 1 & 0 & 1 & 0 & 1 & 0 & 1 & 0  \\
\bottomrule
\end{tabular}
\label{tbl_eta_mnist1}
\end{table}
\renewcommand{\baselinestretch}{2}

\renewcommand{\baselinestretch}{1.2}
\setlength{\tabcolsep}{7pt}
\begin{table}[h]
\centering
\small
\caption{Evolution of condition numbers of stacked weight matrices of a 9-layer neural network with 100 neurons per hidden layer, trained on our MNIST example. Condition numbers indicate that layers 2 through 9 have excessive neurons compared to the first layer, as we expect.}
\begin{tabular}{crrrrrrrrrrr}  
\toprule
\multirow{3}{*}{\bf Layer $(i)$} & \multirow{3}{*}{$n_i$} & \multicolumn{10}{c}{ $\eta$ }  \\ 
\cmidrule(lr){3-12}
& & \multicolumn{2}{c}{.5} & \multicolumn{2}{c}{1} & \multicolumn{2}{c}{2} & \multicolumn{2}{c}{5} & \multicolumn{2}{c}{10} \\ 
\cmidrule(lr){3-12}
& & $\kappa_i$ & $p_i$ & $\kappa_i$ & $p_i$ & $\kappa_i$ & $p_i$ & $\kappa_i$ & $p_i$ & $\kappa_i$ & $p_i$ \\ 
\midrule
 1 & 100 & 5 & 0 & 5 & 0 & 5 & 0 & 5 & 0 & 6 & 0  \\
 2 & 100 & 124 & 4 & 119 & 4 & 144 & 4 & 249 & 4 & 339 & 4  \\
 3 & 100 & 218 & 5 & 245 & 5 & 236 & 5 & 332 & 5 & 398 & 5  \\
 4 & 100 & 101 & 5 & 108 & 5 & 106 & 5 & 109 & 5 & 129 & 5  \\
 5 & 100 & 196 & 5 & 219 & 5 & 241 & 5 & 249 & 5 & 229 & 5  \\
 6 & 100 & 439 & 4 & 424 & 4 & 462 & 4 & 449 & 4 & 332 & 5  \\
 7 & 100 & 184 & 4 & 182 & 4 & 176 & 4 & 166 & 4 & 151 & 5  \\
 8 & 100 & 365 & 5 & 283 & 5 & 256 & 5 & 222 & 5 & 278 & 5  \\
 9 & 100 & 153 & 4 & 180 & 5 & 188 & 5 & 213 & 5 & 229 & 5  \\
 10 & 10 & 2 & 0 & 2 & 0 & 2 & 0 & 2 & 0 & 2 & 0  \\
\bottomrule
\end{tabular}
\label{tbl_eta_mnist2}
\end{table}
\renewcommand{\baselinestretch}{2}

\subsubsection*{Choosing $\tau$ and consistency of results} In our experience, different choices of $\tau$ within a reasonable range (25 - 40) \dd{do} not affect the final outcome (See Table \ref{tbl_sq_mnist1}). 
As mentioned earlier, for any network, the condition numbers of its stacked weight matrices and their variance among the layers of the network can be the best indicator of redundancies present in the network. When some of the layers have condition numbers much larger than others, one could conclude that those layers have excessive neurons compared to others. On the other hand, when condition numbers have small value and small variance, it indicates that the neurons are well-distributed among the layers. Such a network might still need to be scaled up or down, to achieve the best accuracy.

Looking at the mean and variance of the condition numbers \dd{guides} us to choose a good value for $\tau$ for various datasets. If unsure about choosing the $\tau$, the mean of condition numbers among the layers \dd{is a reasonable choice}. 

Table \ref{tbl_tau_mnist1} shows the evolution of a 9-layer network, when processed by Algorithm \ref{alg_1} with $\tau = 30$. It only takes 7 iterations until the network satisfies $\tau \leq 30$ for all of its layers. Performing these 7 iterations take less than 3 minutes on a 2017 Macbook.

\renewcommand{\baselinestretch}{1.2}
\setlength{\tabcolsep}{5pt}
\begin{table}[h]
\centering
\small
\caption{Evolution of number of neurons $n_i$ for a 9-layer neural network, trained on our MNIST example, as it is refined with Algorithm \ref{alg_1}.}
\begin{tabular}{ccccccccc}  
\toprule
\multirow{2}{*}{\bf Layer $(i)$} & \multicolumn{8}{c}{ $n_i$ for iterations of Algorithm \ref{alg_1} }  \\ 
\cmidrule(lr){2-9}
& 0 & 1 & 2 & 3 & 4 & 5 & 6 & 7  \\ 
\midrule
1 & 600 & 600 & 600 & 600 & 600 & 600 & 600 & 600  \\
2 & 540 & 536 & 533 & 532 & 530 & 529 & 529 & 527   \\
3 & 480 & 480 & 477 & 474 & 471 & 470 & 468 & 466   \\
4 & 420 & 420 & 419 & 418 & 418 & 415 & 413 & 412   \\
5 & 360 & 360 & 360 & 360 & 360 & 360 & 359 & 359   \\
6 & 300 & 300 & 300 & 300 & 300 & 300 & 300 & 300   \\
7 & 240 & 240 & 240 & 240 & 240 & 240 & 240 & 240   \\
8 & 180 & 180 & 180 & 180 & 180 & 180 & 180 & 180   \\
9 & 120 & 120 & 120 & 120 & 120 & 120 & 120 & 120   \\
10 & 10 & 10 & 10 & 10 & 10 & 10 & 10 & 10  \\
\midrule
max($\kappa_i$) & 49.4 & 42.5 & 39.2 & 33.6 & 31.7 & 31.4 & 31.0 & 29.2 \\
\midrule
$\Sigma (n_i)$ & 3,250 & 3,246 & 3,239 & 3,234 & 3,229 & 3,224 & 3,219 & 3,214  \\
\bottomrule
\end{tabular}
\label{tbl_tau_mnist1}
\end{table}
\renewcommand{\baselinestretch}{2}

\dr{What does "adaptive" mean in the next sentence?} \olive{I meant $\tau$ is the mean of condition numbers obtained at each iteration. I changed the sentence.}

When we apply Algorithm \ref{alg_1} to the same network, \ds{choosing $\tau$ at each iteration to be the mean of condition numbers,}
it takes 6 iterations to achieve a very similar network. 

Our algorithms are also not very sensitive to the choice of network that we start with. This is mainly because we have decoupled the question of finding the right proportion of neurons for the layers of the network (Algorithm \ref{alg_1}), and the question of finding the overall size of the network (Algorithm \ref{alg_2}). Still, being smart in choosing the initial network structure can significantly reduce the time it takes for the algorithms to refine the structure.

For example, consider the example in Table \ref{tbl_tau_mnist1}. Instead of starting from the network shown for iteration 0 of Table \ref{tbl_tau_mnist1}, we start with a 9-layer network that has 600 neurons on all of its hidden layers. This time, it takes 47 iterations for Algorithm \ref{alg_1} to adjust the distribution of neurons, way more than 7 iterations. However, the output is very similar as shown in Table \ref{tbl_tau_mnist2}, considering the total number of neurons which are 3,214 and 3,298 for the two networks, and the number of trainable parameters which are 1,270,596 and 1,266,728, respectively. 

\renewcommand{\baselinestretch}{1.2}
\setlength{\tabcolsep}{5pt}
\begin{table}[h]
\centering
\small
\caption{Different starting networks lead to similar networks in our experiments. The output of Algorithm \ref{alg_1} for a network with 600 neurons on all its hidden layers, is very similar to the output of Algorithm for a different network in Table \ref{tbl_tau_mnist1}.}
\begin{tabular}{ccc}  
\toprule
\multirow{2}{*}{\bf Layer $(i)$} & \multicolumn{2}{c}{ $n_i$ }  \\ 
\cmidrule(lr){2-3}
 & Input to Algorithm \ref{alg_1} & Output of Algorithm \ref{alg_1} \\ 
\midrule
1 & 600 & 600   \\
2 & 600 & 518  \\
3 & 600 & 448  \\
4 & 600 & 389 \\
5 & 600 & 344  \\
6 & 600 & 300  \\
7 & 600 & 267 \\
8 & 600 & 238  \\
9 & 600 & 184 \\
10 & 10 & 10 \\
\midrule
$\Sigma (n_i)$ & 5,410 & 3,298 \\
\bottomrule
\end{tabular}
\label{tbl_tau_mnist2}
\end{table}

Finally, we note that the final networks \olive{can be slightly different depending} on the starting network, since there are generally many networks that fit the training data. Overall, in our experience, Algorithm \ref{alg_2} chooses a similarly sized scaled-up network, if we start with a smaller network for Algorithm \ref{alg_1}. The key point is obtaining networks of similar accuracy and size, not obtaining a particular network. 
\olive{Clearly, using an oversized network as the input to Algorithm \ref{alg_1} and then contacting it with Algorithm \ref{alg_2} will be computationally more expensive than the alternative approach of starting with a modest network for \ref{alg_1} and then expanding it with Algorithm \ref{alg_2}. This is because, in the latter case, the partial training of networks by Algorithm \ref{alg_1} will be performed on a smaller network.
}	

\section{Conclusion} \label{sec_conclusion}

We have defined a complete set of \olive{inexpensive and approachable} tools that can be used to design a feed-forward neural network from scratch, given only the number of layers that should be used. Although additional computations are used to refine the number of neurons, these computations are overall much less expensive than the alternative method of training many models and choosing the one with best accuracy. Results on sample image and non-image datasets demonstrate that our method results in small networks with high accuracies. By choosing the number of neurons wisely, we avoid both over-fitting and under-fitting of the data and therefore, achieve low generalization errors.
This enables practitioners to effectively utilize compact neural networks for real-world applications. \olive{We also provided a straightforward method for squeezing networks that are already trained. Our method identifies and discards redundancies in the trained networks, leading to compact networks, sometimes with better accuracies.}

\bibliographystyle{siamplain}
\bibliography{references}
\end{document}